\documentclass[dvipsnames]{article} 
\usepackage{iclr2024_conference}
\usepackage{times}

\usepackage{hyperref}
\usepackage{url}
\usepackage[utf8]{inputenc} 
\usepackage[T1]{fontenc}    
\usepackage{hyperref}       
\usepackage{url}            
\usepackage{booktabs}       
\usepackage{amsfonts}       
\usepackage{nicefrac}       
\usepackage{microtype}      
\usepackage{xcolor}         
\usepackage{todonotes}
\usepackage{booktabs}
\usepackage{multirow}
\usepackage{paralist}
\usepackage{tabularx}
\usepackage{enumitem}

\usepackage{latexsym}
\usepackage{paralist}
\usepackage{xspace}
\usepackage[normalem]{ulem}
\useunder{\uline}{\ul}{}
\usepackage{tabu}
\usepackage{tabularx}
\usepackage{makecell}
\usepackage{multirow}
\usepackage{ulem}
\usepackage{algorithm}
\usepackage{algorithmic}
\usepackage{wrapfig}
\usepackage{verbatim}

\usepackage{amsmath,amssymb,bm,dsfont}

\def\1{\bm{1}}

\def\vs{{\bm{s}}}

\DeclareMathAlphabet{\mathsfit}{\encodingdefault}{\sfdefault}{m}{sl}
\SetMathAlphabet{\mathsfit}{bold}{\encodingdefault}{\sfdefault}{bx}{n}

\usepackage{xspace}
\makeatletter
\DeclareRobustCommand\onedot{\futurelet\@let@token\@onedot}
\def\@onedot{\ifx\@let@token.\else.\null\fi\xspace}

\def\eg{\textit{e.g}\onedot} 
\def\ie{\textit{i.e}\onedot} 
 
 \def\vs{\textit{vs}\onedot}
\def\wrt{\textit{w.r.t}\onedot} 

\makeatother

\usepackage{color, colortbl}
\definecolor{emerald}{rgb}{0.31, 0.78, 0.47}
\definecolor{Gray}{gray}{0.9}
\definecolor{Highlight}{rgb}{0.89,0.89,0.94}
\usepackage[first=0,last=9]{lcg}
\newcommand{\chl}{\cellcolor{Highlight}}

\newcommand{\textbi}[1]{\textbf{\textit{#1}}}

\usepackage{pifont}
\newcommand{\xmark}{\ding{55}}%
\newcommand{\cmark}{\ding{51}}%

\newcommand{\Sec}{§}
\usepackage{hyperref}
\renewcommand{\bm}[1]{\mathbf{#1}}

\newcommand{\method}{\textsc{Diffusion-LLM}\xspace}

\usepackage{cleveref}
\newcommand\myshade{85}
\colorlet{mylinkcolor}{YellowOrange}
\colorlet{mycitecolor}{violet}
\colorlet{myurlcolor}{Aquamarine}
\hypersetup{
  linkcolor  = mylinkcolor!\myshade!black,
  citecolor  = mycitecolor!\myshade!black,
  urlcolor   = myurlcolor!\myshade!black,
  colorlinks = true,
}
\usepackage{titlesec}
\titlespacing{\paragraph}{0pt}{-1pt}{\parskip}
\titlespacing{\section}{0pt}{\parskip}{\parskip}
\titlespacing{\subsection}{0pt}{\parskip}{\parskip}
\titlespacing{\subsubsection}{0pt}{\parskip}{\parskip}

\title{Diffusion Language Models Can Perform Many Tasks with Scaling and Instruction-Finetuning}

\iclrfinalcopy

\author{Jiasheng Ye$^{1,2}$\thanks{Work was done during Jiasheng's internship at ByteDance Research.},~ Zaixiang Zheng$^{\dagger 1}$, Yu Bao$^{1}$, Lihua Qian$^{1}$, Quanquan Gu$^{\ddagger1}$ \\
$^{1}$ByteDance Research~ $^{2}$Fudan University \\
\texttt{jsye23@m.fudan.edu.cn~\{zhengzaixiang,quanquan.gu\}@bytedance.com} \\
{\small $^\dagger$Project Lead ~~$^\ddagger$Corresponding Author} \\
{\small Code available at: \url{https://github.com/yegcjs/DiffusionLLM}}\\
}

\begin{document}

\maketitle

\begin{abstract}

The recent surge of generative AI has been fueled by the generative power of diffusion probabilistic models and the scalable capabilities of large language models~(LLMs).
Despite their potential, it remains elusive whether \method can solve general language tasks comparable to their autoregressive counterparts.
This paper demonstrates that scaling masked discrete diffusion models \wrt data, sizes, and tasks can effectively make them strong language learners.
We introduce {\method}s at scale by first acquiring knowledge from massive data via masked language modeling pre-training thanks to their intrinsic connections.
We then reprogram pre-trained Masked LMs into {\method}s via diffusive adaptation, wherein task-specific finetuning and instruction finetuning are explored to unlock their versatility in solving general language tasks. 
Experiments show that scaling {\method}s consistently improves performance across downstream language tasks. 
We further discover that instruction finetuning can elicit zero-shot and few-shot in-context learning abilities that help tackle many unseen tasks by following both natural language instructions and even visual instructions for multimodal understanding, 
and show promise in advanced and challenging abilities such as reasoning. 
\end{abstract}

\section{Introduction}

Recent advances in generative modeling have led to remarkable progress in the field of generative AI. 
In domains of continuous signals, diffusion probabilistic models have shown great success in rendering photorealistic images~\citep{rombach2021highresolution,ramesh2022dalle2}, immersive videos~\citep{bar2024lumiere} and synthesizing high-quality audio~\citep{kong2020diffwave} through iterative denoising, outperforming GANs and autoregressive (AR) models, even contributing to the surge of AI art. 
The story is different in the domains of discrete signals comprising symbolic sequences such as natural languages, where AR large language models~\citep[large language models or LLMs,][]{brown2020lgpt3,openai2023gpt4} have dominated the scene, delivering impressive generalist language abilities in language understanding and generating human-like texts, and can even follow natural language instructions to perform unseen tasks.

While many recent endeavors try unifying the generation paradigms by enabling large language models to draw~\citep{ge2023seed} or speak~\citep{zhang2023speechgpt}, few explore generating discrete sequences such as languages with diffusion models. 
We suggest that the revolutionized generative abilities of diffusion models give the promise of a strong complement to autoregressive LMs for several favorable reasons, including (1) global receptive field \vs one-sided context, and (2) non-autoregressive drafting-then-revising manner \vs restrictive unidirectional generation/autoregression.
Hence, an intriguing question arises: \textit{can diffusion models speak languages well?}

The key to the great success of modern LLMs lies in their scalability which fosters powerful generalist capabilities.
The question of the capability of {\method}s is thus in turn to ask about their scalability, which can be further boiled down into the following specific research questions regarding the three key ingredients of the success of large-scale LMs, i.e., data, model sizes, and tasks:
\begin{itemize}[itemsep=2pt,topsep=0pt,parsep=1pt,leftmargin=20pt]
\item [(i)] \textit{On scaling data.} 
Acquiring general knowledge via self-supervised pre-training from massive unlabeled data plays a crucial role in the success of the modern NLP paradigms~\citep{radford2018improving,devlin2018bert}, hence it is also of importance to enable Diffusion LMs to learn from massive data.
\textbi{Can {\method}s leverage knowledge from large-scale data?}
\item [(ii)] \textit{On scaling model sizes.} It has been widely observed that the larger the model size, the more competent the LMs become. 
\textbi{Can enlarging 
{\method}s effectively improve downstream tasks?}
\item [(iii)] \textit{On scaling tasks.} 
What makes LLMs most attractive is they can tackle new tasks that they were never exposed to during training by following natural language and even multimodal instructions with little to no demonstrations.
\textbi{Can {\method}s exhibit general zero-shot and few-shot in-context learning capabilities to generalize to unseen tasks?}
\end{itemize}

\begin{figure*}[t]
  \centering
  \includegraphics[width=\linewidth]{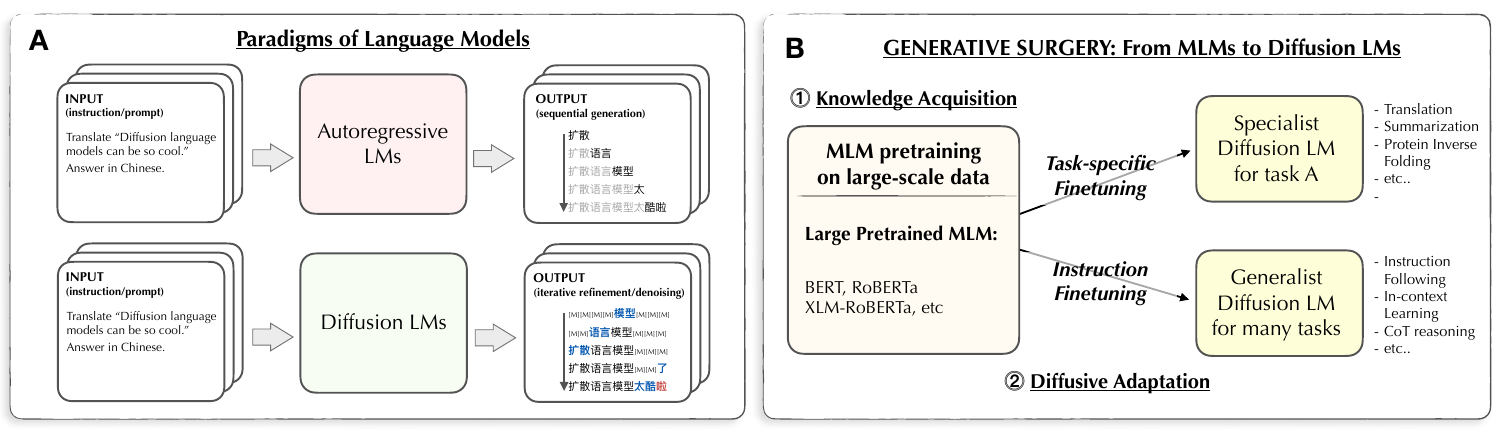}
  \vspace{-2mm}
  \caption{Overview. (A) Comparative illustration of LM paradigms, \ie, autoregressive LMs \vs Diffusion LMs. (B) Overall illustration of the proposed \method where large-scale pre-trained masked LMs are reprogrammed to {\method}s via \textit{generative surgery}.}
  \label{fig: main}
  \vspace{-3mm}
\end{figure*}

In this paper, we delve into the potential of {\method}s through the three research questions.
We highlight our contributions and findings as follows:

\textbf{(1)} We first demonstrate the intrinsic connection between masked LMs and discrete diffusion models, which permits us to treat pre-trained masked LMs of various scales as pre-trained {\method}s, without the need for expensive learning from scratch. 
We then reprogram pre-trained masked LMs into {\method}s via \textit{diffusive adaptation}, where task-specific finetuning and instruction finetuning~\citep{wei2021flanv1} are explored for solving certain downstream tasks or general language problems, showing {\method}s benefit from pre-training on large scale data.  
\textbf{(2)} 
We reveal that large-scale {\method}s can serve as strong sequence generative models to tackle multitasks, exhibiting competitive performance compared with autoregressive LMs.
And the performance consistently improves as the model sizes scale up. 
\textbf{(3)} We further elicit zero-shot and few-shot abilities for {\method}s to tackle multiple unseen tasks, spanning from language and visual ones, through both language and vision instruction finetuning.
Notably, {\method}s demonstrate promising structured reasoning behaviors thanks to their flexible non-autoregressive generation order. Nevertheless, their capacity to tackle complex reasoning tasks remains an ongoing challenge awaiting resolution.



To sum up, we hope that our explorations provide valuable insights into the scalability of {\method}s and their potential as a viable complement in tackling generative language tasks across the board.


\section{Preliminaries: Diffusion Models for Sequence Generation}
\label{sec: preliminary}

Language processing tasks can be unified as sequence-to-sequence problems~\citep{raffel2020T5}, modeling the conditional distribution $p_\theta(\bm{x}| \bm{c})$, where $\bm{x}=(\bm{x}^{[1]}, \bm{x}^{[2]}, \dots, \bm{x}^{[N]})$ is a target sequence composing $N$ tokens and $\bm{c}$ is the given context.
For example, we may want to generate responses $\bm{x}$ conditioned on the prompt $\bm{c}$, or it can be unconditional generation if no context is provided (\ie, $\bm{c}=\phi$). 
As a result, one thing we care about is the capability of generative models for sequence data $\bm{x}$, \eg, the prevailing autoregressive models or diffusion models.
In this section, we provide the necessary background on diffusion-based sequence generative models, where we abuse the notation and use $p_\theta(\bm{x})$ for both conditional $p_\theta(\bm{x} | \bm{c})$ and unconditional $p_\theta(\bm{x} | \bm{c}=\phi)$ for brevity.

\paragraph{Diffusion Models} \citep{diffusion2015} are a class of generative models characterized by a pair of Markov processes, \ie, a forward diffusion process and a backward denoising process. 
The \textit{forward} process~${q(\bm{x}_{1:T}|\bm{x}_0)=\prod_{t=1}^T q(\bm{x}_t|\bm{x}_{t-1})}$ gradually perturb the data $\bm{x}_0\sim q(\bm{x}_0)$ into a stationary distribution $q(\bm{x}_T)$ with $T$ increasingly noisy steps $\bm{x}_{1:T}=\bm{x}_{1}, \bm{x}_2, \dots, \bm{x}_T$. 
The learned \textit{backward} process ${p_{\bm{\theta}}(\bm{x}_{0:T}) = p(\bm{x}_T)\prod_{t=1}^{T}p_{\bm{\theta}}(\bm{x}_{t-1}|\bm{x}_t)}$, reversely, gradually denoises the samples towards the data distribution. 
To fit the model $p_{\bm{\theta}}(\bm{x}_0)$ to the data distribution $q(\bm{x}_0)$, the denoiser model is typically optimized by the variational bound of the negative log-likelihood~\citep{ddpm}:
\begin{align}
\setlength{\abovedisplayskip}{0.25pt}
\setlength{\belowdisplayskip}{0.25pt}
      \mathbb{E}_{q(\bm{x}_0)}\left[-\log p_\theta(\bm{x}_0)\right] \le \mathbb{E}_{q(\bm{x}_{0:T})} \left[-\log \frac{p_{\bm{\theta}}(\bm{x}_{0:T})}{q(\bm{x}_{1:T}|\bm{x}_0)}\right] 
      = \mathcal{L}_0 + \sum_{t=2}^T \mathcal{L}_t+ \text{const.},
      \label{eqn: variational bound}
\end{align}
where $\mathcal{L}_0=\mathbb{E}_{q}\left[-\log p_{\theta}(\bm{x}_0 | \bm{x}_1)\right]$, and $\mathcal{L}_t=\mathbb{E}_{q}\left[\text{KL}[q(\bm{x}_{t-1}|\bm{x}_t, \bm{x}_0)\|p_{\bm{\theta}}(\bm{x}_{t-1}|\bm{x}_t)]\right]$ for $t \in [1, T]$.  


In general, diffusion models can be categorized into continuous and discrete diffusion models according to distribution type for data perturbation.
Continuous diffusion models with Gaussian perturbation have demonstrated impressive performance in generating continuous signals~\citep{rombach2021highresolution, ho2022imagenvideo, kong2020diffwave} but still struggle with satisfactory generation quality in natural languages~\citep{diffusionlm,gong2022diffuseq,gao2022difformer,yuan2022seqdiffuseq,ye2023dinoiser}.
A critical challenge herein is the \textit{pitfall of discreteness}~\citep{ye2023dinoiser} that makes Gaussian perturbation on embeddings hardly provide effective training signals.
In contrast, discrete diffusion models directly operate over the discrete state space of tokens, providing an attractive alternative for generative sequence learning. Therefore in this paper, we explore developing {\method}s upon discrete diffusion.


\paragraph{Discrete Diffusion Models}~\citep{hoogeboom2021argmax,austin2021structured} cover a subset of diffusion models for which transition probabilities between timesteps are discrete distributions. 
Since the forward diffusion process is applied independently to each token of a sequence $\bm{x}$, for the sake of brevity, we abuse the notation $\bm{x}_t$ for arbitrary tokens at diffusion timestep $t$.
Formally, $\bm{x}_t\in\{0, 1\}^{|\mathcal{V}|}$ is a token represented as a one-hot vector, where $\mathcal{V}$ is the vocabulary of all possible tokens.
Let $\texttt{Cat}(\bm{x};\bm{p})$ be a categorical distribution on $\bm{x}$ with probabilities given by vector $\bm{p}$ on $|\mathcal{V}|-1$ dimensional probability simplex, and the forward transition be $q(\bm{x}_t|\bm{x}_{t-1})=\texttt{Cat}\left(\bm{x}_t; \bm{p}=\beta_t\bm{x}_{t-1} + (1-\beta_t)\bm{q}_{\text{noise}}\right),$
where $0\ll\beta_t<1$ is the noise schedule controlling the degree of perturbation at timestep $t$, and $\bm{q}_\text{noise}$ is the probability vector of stationary distribution $q(\bm{x}_T)$, \ie, $q(\bm{x}_T)=\texttt{Cat}(\bm{x}_T; \bm{p}=\bm{q}_{\text{noise}})$. 
In this case, the distribution of corrupted sample $\bm{x}_t$ given its original data $\bm{x}_0$ has a closed-form expression:
\begin{equation}
\nonumber
q(\bm{x}_t|\bm{x}_{0})= \texttt{Cat}\left(\bm{x}_t; \bm{p}=\alpha_t\bm{x}_0+ (1-\alpha_t)\bm{q}_{\text{noise}}\right),
\label{eqn: dd 0_to_t}
\end{equation}
where $\alpha_t = \prod_{i=1}^t\beta_i$. This shows that the diffusion process is intuitively a convex combination between data and noise where the $\alpha_t$ controls the degree of corruption at different timesteps. In particular, $\alpha_t$ decreases as the timestep increases. With sufficiently large timesteps, we have $\alpha_T\approx 0$, which preserves no information from the data at the end of the diffusion process. 

Different stationary distributions $\bm{q}_\text{noise}$ lead to different formulations of discrete diffusion models. One typical design is the \textit{absorbing} diffusion with $q(\bm{x}_T) = \{ 1~\text{if}~ \bm{x}_T = \texttt{[MASK]};~0 ~\text{if}~ \bm{x}_T \not= \texttt{[MASK]} \}$, where \texttt{[MASK]} is an absorbing state. 
According to Eq.~\eqref{eqn: dd 0_to_t}, this formulation results in $\bm{x}_t$ either being masked or the same as $\bm{x}_0$, with a masking ratio $(1-\alpha_t)$. 
This makes absorbing diffusion resemble masked LMs~\citep[MLM,][]{devlin2018bert} as \citet{he2022diffusionbert} points out.

\paragraph{Reparameterized Discrete Diffusion Models}~\citep[RDM,][]{zheng2023reparameterized} reparameterize the backward transition of {\method}s that reformulates the training objective of discrete diffusion models into 
\begin{align}
\mathcal{L}_t = \mathbb{E}\big[-\lambda_{t-1}^{(2)}\left(1-\mathds{1}(\bm{x}_t=\bm{x}_0)\right)\log p_{\bm{\theta}}(\bm{x}_0|\bm{x}_t)\big], 
\label{eqn: reparam objective}
\end{align}
where $\mathds{1}(\cdot)$ is indicator function. 
Under the formulation of absorbing diffusion, Eqn.~\ref{eqn: reparam objective} resembles a weighted MLM objective~\citep{devlin2018bert}.
\citet{zheng2023reparameterized} demonstrate that Eqn.~\ref{eqn: reparam objective} is a more effective training protocol compared to Eqn.~\ref{eqn: variational bound} for generative discrete diffusion models, showing performance on par with autoregressive LMs~\citep{vaswani2017attention} 
on representative machine translation benchmarks for the first time. 
In this paper, we use RDM as our primary training objective for building our {\method}s (see \S\ref{app: details} for more details).

\paragraph{Generative Process of Discrete Diffusion Models.} 
Diffusion models yield new samples by their reverse generative process of iterative denoising.
Under the formulation of absorbing diffusion, the denoising process can be characterized in an iterative \textit{mask-predict} manner~\citep{ghazvininejad2019mask}. 
Specifically, the starting sequence is initialized by all $\texttt{[MASK]}$ tokens, and in each iteration, some masked tokens are replaced by the model predictions from $p_{\theta}(\bm{x}_{t-1}|\bm{x}_t)$ while some unmasked tokens are remasked, according to specific strategies/schedules~\citep{ghazvininejad2019mask,savinov2021step,chang2022maskgit,zheng2023reparameterized}.
In this paper, we follow \citet{zheng2023reparameterized} to unmask positions with top-$k$ predicted $\log p_{\theta}(\bm{x}_{0}|\bm{x}_t)$, and mask all the rest position in each denoising step\footnote{See \S\ref{app: details} for concrete noise schedules, and \citet{zheng2023reparameterized} for the justification of this sampling strategy.}.

\section{Scaling Diffusion Language Models \wrt Data, Sizes and Tasks}


Developing {\method}s that leverage the advantages of both the generative power of both diffusion models and the scalability of large pre-trained LMs is a promising yet challenging endeavor.
The key to the success of the current standard paradigm of large generative LMs is acquiring knowledge via massive pre-training and generating in a prompt-response manner for preferable output for many tasks. 
For {\method}s, (1) how to benefit from pre-training at scale, and (2) how to best fit the prompt-response paradigm, are the crucial open questions. 
In this section, we will elaborate on how to empower {\method}s with knowledge from pre-training of large-scale data as well as model sizes, and extend their generative capabilities for extensive downstream tasks.

\subsection{Knowledge Acquisition via MLM pre-training}
\label{sec: pre-train}
The theoretical framework of discrete diffusion models has an intrinsic connection to masked language modeling (MLM), which was discussed in~\cite{austin2021structured,gong2022diffuseq} and \citet{he2022diffusionbert}.
Among various types of discrete diffusion models, the \textit{absorbing} diffusion~\citep{austin2021structured} resembles a \textit{generalized} masked language modeling, which has been shown to be an effective training objective in pre-training foundation models~\citep{devlin2018bert,liu2019roberta}.
Specifically, absorbing diffusion defines a stationary distribution: $q(\bm{x}_T) = \{ 1~\text{if}~ \bm{x}_T = \texttt{[MASK]};~0 ~\text{if}~ \bm{x}_T \not= \texttt{[MASK]} \} $,
where \texttt{[MASK]} is an absorbing token. 
According to Eq.~\eqref{eqn: dd 0_to_t}, this formulation results in $\bm{x}_t$ either being masked or the same as $\bm{x}_0$, with a masking ratio $(1-\alpha_t)$. 
Consequently, $\bm{x}_t=\bm{x}_0$ if and only if $\bm{x}_t\not=\texttt{[MASK]}$, which aligns the reparameterized training objective in Eq.~\eqref{eqn: reparam objective}  exactly with the masked language modeling objective. 

This connection allows us to establish {\method}s by pre-training with MLM objectives from massive raw textual data. 
We can even treat abundant community-available pre-trained MLMs~\citep{devlin2018bert,liu2019roberta,conneau2019xlmr} as pre-trained {\method}s, and can depart from them for downstream tasks at a very low cost, bypassing the expensive pre-training stage.



\subsection{Diffusive Adaptation: Reprogramming pre-trained MLMs to {\method}s}

Existing masked LMs are primarily designed to serve as sequence encoders, and are not able to generate sequences by default.
Despite their connections to absorbing discrete diffusion, it is non-trivial to naively sample from masked LMs through the iterative denoising process of absorbing diffusion. 
One major reason is that absorbing diffusion generates sampling by iterative applying $p_{\theta}(\bm{x}_{t-1}|\bm{x}_t)$ from complete noise to the final prediction (\ie, ranging gradually from $100\%$ to $0\% \texttt{[MASK]}$ tokens) through different timesteps, whereas vanilla masked LMs are only pre-trained with a limited and constant masking ratio (\eg, 15\%). 

In order to elicit the pre-trained masked LMs' ability for sequence generation, we propose \textit{diffusion adaptation} to eliminate the gap between pre-trained masked and {\method}s, where we further finetune pre-trained Masked LMs with diffusion training objective such that sampling with the denoising process becomes possible.
In particular, we follow the reparameterized training and sampling method in~\citet{zheng2023reparameterized} as described in \S\ref{sec: preliminary}.
Similar to the time agnostic design in \citet{he2022diffusionbert}, we do not introduce any extra parameters to differentiate different diffusion timesteps. 

For different purposes, we perform diffusive adaptation for {\method}s in two ways:
\begin{itemize}[itemsep=2pt,topsep=0pt,parsep=1pt,leftmargin=10pt]
\item \textbi{Optimizing specialist capabilities on certain downstream tasks via task-specific finetuning.} 
To verify the feasibility of diffusive adaptation, we finetune pre-trained masked LMs on specific datasets for each downstream task.
Moreover, we further perform finetuning on pre-trained models of different scales so as to study the scalability of {\method}s.

\item \textbi{Eliciting generalist capabilities on extensive tasks via instruction finetuning.}
Finetuning on a collection of tasks phrased as instructions (\ie, instruction finetuning) enables LMs to better respond to instruction prompts and generalize to unseen tasks~\citep{wei2021flanv1,chung2022scalingflan}.
Inspired by this, we apply diffusive adaptation to pre-trained masked LMs by instruction finetuning to study whether {\method}s can acquire few-shot and zero-shot abilities like autoregressive LLMs.
\end{itemize}

\paragraph{Engineering considerations for scaling.}
Both scenarios above handle conditional sequence generation tasks from input to output, which require the model to generate target sequences according to the given prompts.
Instead of incorporating an extra encoder,
we handle conditional generation by organizing data in a prompt-response format\footnote{
A prompt-\underline{response} formatted example for German$\to$English translation (Vielen dank$\to$Thank you):
``Translate the German sentence to English. German: Vielen dank. English: \underline{Thank you.}''
}.
This enables our model to share the same training infrastructure as prevailing large decoder-only AR LMs~\citep{brown2020lgpt3,touvron2023llama,touvron2023llama2,jiang2023mistral} while incorporating extra encoder adds to the complexity in key techniques for scaling LLMs such as pipeline parallelism~\citep{harlap2018pipedream,huang2019gpipe,li2021chimera}.
Our design decision ensures the scalability of our {\method}s in engineering.

Besides, we incorporate a length predictor, a common practice in non-autoregressive text generation~\citep{gu2018non}, to determine the lengths of predicted sequences. 
We pick its top-$k$ length predictions for parallel length beam search, where $k$ is referred to as the length beam size.
During tuning, we only apply the diffusion process to the target response tokens and compute loss on them. 
During inference, we append the initial fully masked sequences to the prompts and denoise from them. 

\section{Experiments}
\label{sec: exp setup}

In this section, we first introduce our general experimental setups in \S\ref{sec: exp setup}. Then we conduct three parts of experiments progressively regarding scaling on data (\Sec\ref{sec: finetune}), model sizes (\Sec\ref{sec: scaling}), and the number of tasks (\Sec\ref{sec: instruction tuning}).

\begin{table*}[t]
\centering
\setlength{\tabcolsep}{1.7pt}
\caption{
SacreBLEU~\citep{post2018call} on IWSLT14 \textsc{De}$\rightarrow$\textsc{En} and WMT14 \textsc{En}$\rightarrow$\textsc{De}, and Rouge-L on Gigaword-10k. 
We use 10 length beams for all the results with length prediction.
Results out of (inside) parentheses are obtained with length prediction (oracle target length).
Results of DiffusionLM and \textsc{DiNoiser} are quoted from \citet{ye2023dinoiser}.
And the results of GENIE are our reimplementation obtained with the open-source code and checkpoint of the orignal paper.
``\#Params.'': Number of non-embedding parameters. 
``Type'': whether the training objective and sampling method are autoregressive~\citep[AR,][]{vaswani2017attention} or follow reparameterized diffusion models~\citep[RDM,][]{zheng2023reparameterized}.
``pre-trained'': whether initialized from pre-trained models.
``$\dag$'': The architecture is in the format of \# layers / hidden dimension. For models w/ encoder, the \# layers consist of layers in both encoder and decoder.
``$\ddag$'': For IWSLT14, we follow previous practice~\citep{vaswani2017attention} and use a smaller version (39M) of \texttt{Transformer-BASE}, which has a dimension of 1024 in the feed-forward layers is 1024 and 4 attention heads. 
}
\label{tab: mt-finetune}
\resizebox{\linewidth}{!}{
\small
\begin{tabular}{llcccrrrr}
\toprule
                            & Method & Architecture$^\dag$  & \#Params.  & Pre-trained & IWSLT14 & WMT14 & Gigaword-10K & Gigaword \\ 
\midrule
\multirow{5}{*}{w/ encoder} & AR~\citep{vaswani2017attention}& 12/512  &  39-43M$^\ddag$ & \xmark &33.30&26.85&10.42&34.5\\
                            &DiffusionLM~\citep{diffusionlm} & 12/512  & 39-43M$^\ddag$ & \xmark &29.11&17.41&-&-\\
                            & \textsc{DiNoiSer}~\citep{ye2023dinoiser}& 12/512 & 39-43M$^\ddag$& \xmark&31.44&24.62&-&-\\
                            & RDM~\citep{zheng2023reparameterized} & 12/512 & 39-43M$^\ddag$& \xmark         &32.14 &26.54&16.25&- \\
                            \cmidrule[0.4pt]{2-9}
                            & GENIE~\citep{lin2022genie} & 12/768 & 86M  & \cmark         &-     & -   & -   &30.24\\  
\midrule
\multirow{4}{*}{w/o encoder}    & AR~\citep{vaswani2017attention} & 12/768  &  86M  & \xmark         &26.07& -&4.7&- \\
                                & RDM~\citep{zheng2023reparameterized} &12/768&86M           & \xmark         & 28.79 (29.12)&  26.09 (26.86)&10.01 (10.66)&-\\ 
\cmidrule[0.4pt]{2-9}
    & \chl \method (RDM + XLM-R) & \chl 12/768& \chl 86M   & \chl \cmark &\chl 34.10 (35.78) &\chl 26.65 (26.64)& \chl 27.52 (28.83)&\chl34.08(35.65)\\
    & \chl \method (RDM + XLM-R) & \chl 48/4096& \chl 9.7B  & \chl \cmark &\chl \bf 38.57 (40.65) &\chl \bf 30.34 (32.81)& \chl \bf 31.54 (32.57)&\chl \bf 34.24(34.26)\\
\bottomrule
\end{tabular}
}
\end{table*}

\paragraph{Model.} 
Throughout the experiments, we use XLM-RoBERTa~\citep[\texttt{XLM-R};][]{conneau2019xlmr,goyal2021xlmr_xl} as our foundation models, which is pre-trained on CC100~\citep{wenzek2020ccnet}, a multilingual corpus containing 167B tokens of 100 languages, with four model sizes (numbers of non-embedding parameters) at different scales, \ie, 86M, 304M, 2.8B, and 9.7B.

\paragraph{Data \& task.} We investigate our approach for its specialist ability in respective downstream tasks and generalist ability to solve massive unseen tasks using natural language instructions. The datasets we use to finetune our model are as follows:

\textbi{(1) Downstream task datasets.} 
We evaluate whether our approach can help {\method}s serve as strong specialized models on multiple representative downstream tasks: 
(1)~IWSLT14 for \textsc{De}$\rightarrow$\textsc{En} translation; (2) WMT14 for \textsc{En}$\rightarrow$\textsc{De} translation; (3) Gigaword-10K for text summarization; and (4) the full Gigaword summarization. 

\textbi{(2) Instruction finetuning datasets.}
We follow \citet{chung2022scalingflan} and finetuned XLM-R models of different scales with the Flan 2022 Collection~\citep{chung2022scalingflan,longpre2023flan} with diffusion training objective.
It is the publicly available version of the instruction tuning data for Flan-T5 and Flan-PaLM, covering over 1.8K tasks. 
It combines several multitask learning datasets with instructions~\citep{wei2021flanv1,sanh2021multitask,supernaturalinstructions}, combined with a few extra chain-of-thought and dialog data.

\begin{figure*}[t]
  \centering
  \vspace{-1mm}
  \includegraphics[width=1.0\linewidth]{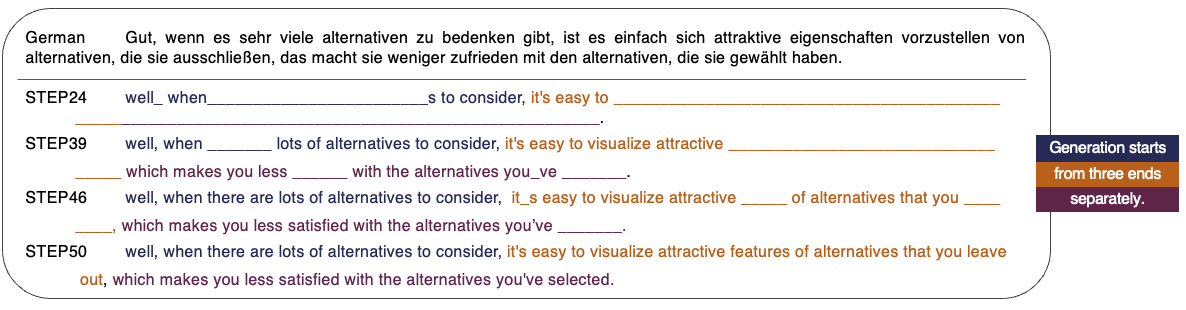}
  \vspace{-6mm}
  \caption{An exemplary generation process of \method for machine translation. Notice that the target translation contains three segments, which are generated simultaneously by the diffusion language model.}
  \label{fig: qualitative mt}
\end{figure*}

\subsection{Exploring the Design Space of {\method}s}

We first validate that our design decisions lead to a competitive {\method}, which includes (1) time-agnostic RDM and (2) no incorporation of an extra encoder. As shown in Tab.~\ref{tab: mt-finetune}:
\begin{itemize}[itemsep=2pt,topsep=0pt,parsep=1pt,leftmargin=10pt]
    \item \textbf{RDM is the most performant {\method}, being an ideal candidate for scaling.} 
    It outperforms both continuous diffusion language models, DiffusionLM and \textsc{DiNoiSer}. 
    Moreover, it performs on par with AR model on IWSLT14 and WMT14, outperforming AR on Gigaword-10K, indicating that RDM is a strong enough diffusion model to exploit. 
    \item \textbf{Using an architecture without encoders has a minor effect on model performance.} 
    Despite that model without encoders slightly underperforms their counterpart with encoders, bidirectional perception of diffusion models and scaling in data can mitigate this gap, similar to the findings in \citet{zhang2022examining,patel2022bidirectional}.
    As results in Tab.~\ref{tab: mt-finetune} show, compared to AR models, RDM has a smaller performance gap between models with and without encoders.
    Such discrepancy becomes neglectable on large datasets like WMT14 and models initialized from the pre-trained XLM-R models.
\end{itemize}
The results verify the competence of our model at moderate scales without pre-training, preparing it to be an ideal candidate for scaling.
We then build upon this to investigate its scalability \wrt, data, model sizes, and tasks.

\vspace{-1mm}
\subsection{Empowering {\method}s with Large-scale Data}
\vspace{-1mm}
\label{sec: finetune}
We apply diffusive adaptation on \texttt{XLM-R-BASE}~\citep{conneau2019xlmr} model architecture on sequence generation benchmarks.
In this way, we verify the feasibility of diffusive adaptation.
Meanwhile, by comparing the performance of diffusive adapted RDMs to models trained from scratch, we confirm that our {\method} takes advantage of large-scale self-supervised learning, namely the scaling with data.

\paragraph{pre-training at scale benefit {\method}s. }  The results in Tab.~\ref{tab: mt-finetune} demonstrate that Training RDM through diffusive adaptation from pre-trained MLMs boosts the model performance compared to the models trained from scratch, whose impact is particularly significant on small datasets like Gigaword-10K. 
Compared to GENIE, a pre-trained {\method} based on continuous diffusion models, RDM adapted from XLM-R is also stronger, further confirming the advantage of discrete {\method}. 
As qualitatively shown in Fig.~\ref{fig: qualitative mt}, {\method}s generate fluent and semantically accurate translation\footnote{The intermediate steps demonstrate that the models generate three clauses simultaneously, implying a global perception that plans the generation of the whole sequence.
We consider this benefits the model on more complex generation tasks, which we discuss in \S\ref{sec: qualitative}.}, further confirming the feasibility of our generative surgery to the pre-trained MLMs.
With these findings, we confirm the feasibility of our diffusive adaptation method and verify the scalability of our {\method}s regarding training data.

\begin{figure*}[t]
\centering
\vspace{-2mm}
  \includegraphics[width=\linewidth]{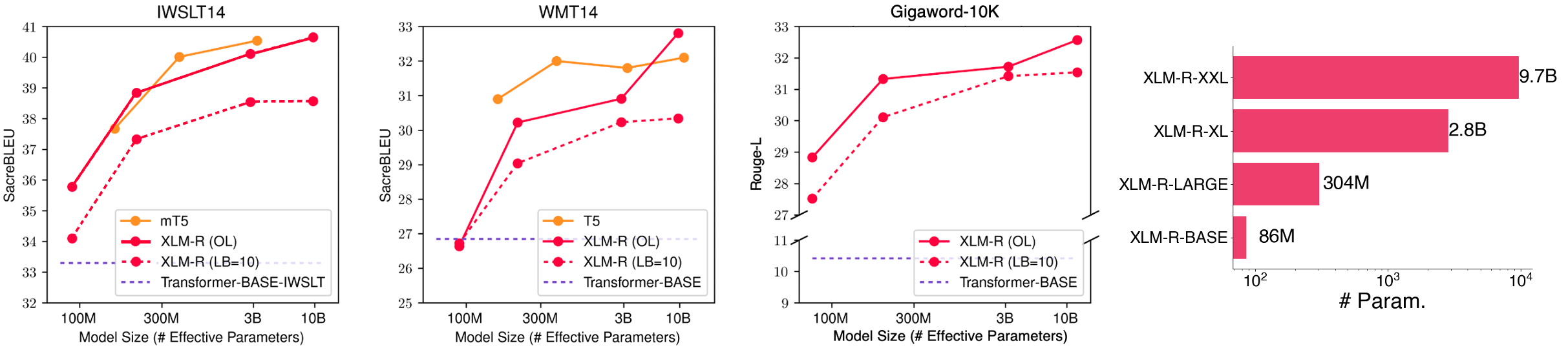}
  \vspace{-6mm}
  \caption{Scaling curves of task-specific finetuning on IWSLT14, WMT14 and Gigaword-10K. We obtain results of mT5~\citep{xue2020mt5} on IWSLT14 by ourselves. The results of T5 on WMT14 are from~\citet{raffel2020T5}. ``OL'': results obtained with oracle target lengths. ``LB=10'': length prediction results with 10 length beams. ``\#Params.'': Number of effective parameters (\ie, non-embedding parameters).}
  \vspace{-3mm}
  \label{fig: scaling}
\end{figure*}

\subsection{Scaling up the Size of {\method}s Boosts Downstream Tasks}
\label{sec: scaling}
We now move on to the scalability with respect to model sizes.
We finetune \texttt{XLM-R} models of different scales~\citep{conneau2019xlmr, goyal2021xlmr_xl}, whose effective parameters (\ie, number of non-embedding parameters) range from $<$100M to 10B. 
Notably, when scaling up to 10B, the model shows impressive performance that surpasses base-sized models by a remarkable margin~(Tab. \ref{tab: mt-finetune}).


Fig.~\ref{fig: scaling} shows the scaling curve of model performance with respect to model sizes.
It demonstrates that the performance of the finetuned diffusion models substantially increases as the model size increases. 
This shows the scaling law of {\method}s in terms of model size.
In addition, we also include the performance of \texttt{(m)T5}~\citep{raffel2020T5,xue2020mt5} at similar scales as references to intuitively understand how scalable our {\method}s are. 
Note that the performance of different models is intricately affected by not only the model size but also numerous factors including model designs, pre-training budget, pre-training objectives, as well as pre-training data~\citep{shazeer2020glu,raffel2020T5,tay2022ul2,scao2022language,hoffmann2022training}.
In Fig.~\ref{fig: scaling}, although we see a performance gap between the finetuned \texttt{(m)T5} and \texttt{XLM-R} models at similar scales, the discrepancy is minor and does not seem amplified as models scale up.
Therefore, while there is still ample room for improving large-scale pre-trained {\method}s, we believe that the path of scaling up these models holds great promise.

\subsection{Instruction-Finetuning Helps Generalize to Unseen Tasks}
\label{sec: instruction tuning}

\begin{wraptable}[12]{r}{0.45\linewidth}
\vspace{-7.5mm}
\centering
\caption{\textbf{Zero-shot} SacreBLEU of instruction-tuned {\method}s on IWSLT14 \textsc{De}$\rightarrow$\textsc{En} translation. 
For Flan 2021, we explicitly remove all German data for strict evaluation. Results are obtained with oracle length.}
\label{tab: flan 2021}
\resizebox{\linewidth}{!}{
\begin{tabular}{lcr}
\toprule
Architecture & Strict Flan'21~ & Flan'22 \\
\midrule
\multicolumn{3}{l}{\underline{\textit{instruction-tuned Diffusion-LLM:}}}      \\
\texttt{XLM-R-BASE} (85M)   & ~~7.15 & 21.26    \\
\texttt{XLM-R-LARGE} (304M)  & 22.52 & 25.24    \\
\texttt{XLM-R-XL} (2.8B)    & 27.27 & 28.13   \\
\texttt{XLM-R-XXL} (9.7B)    & 28.74 & 29.59   \\
\midrule
\multicolumn{3}{r}{ref: \textit{supervised} AR on 160k \textsc{De}$\rightarrow$\textsc{En} data: 33.30} \\
\bottomrule
\end{tabular}
}
\vspace{-3mm}
\end{wraptable}

A fascinating property that motivates scaling LMs up is that LLMs can follow instructions and show impressive few-shot or even zero-shot performance~\citep{wei2021flanv1}.
We now investigate whether diffusion models can also exhibit zero-shot and few-shot performance when being scaled up.

\subsubsection{Instruction finetuning elicits scalable zero-shot performance} 

\begin{figure*}[h]
  \centering
    \includegraphics[width=\linewidth]{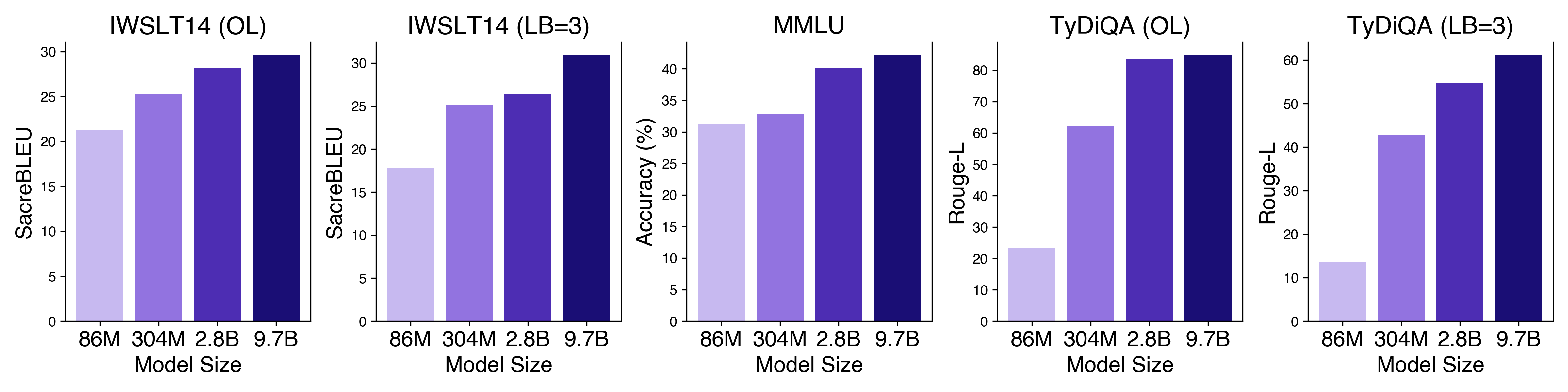}
  \vspace{-4mm}
  \caption{Zero-shot performance of instruction-tuned {\method}s (\texttt{Flan-XLM-R}s) of different scales. OL means the results are obtained with oracle length, while LB means the number of length beams to sample the target with length prediction. The model sizes refer to the number of non-embedding parameters.}
  \label{fig: zero-shot}
\end{figure*}


\paragraph{Strict zero-shot evaluation on IWSLT14 \textsc{De}$\rightarrow$\textsc{En}.}
We first conduct a strict zero-shot evaluation to study if {\method}s can acquire zero-shot capabilities through instruction finetuning.
Specifically, we evaluate on IWSLT14 \textsc{De}$\rightarrow$\textsc{En} translation task, for which we instruction-finetune {\method}s on Flan 2021 Collection~\citep{wei2021flanv1} with all German data removed to ensure that the \textsc{De}$\rightarrow$\textsc{En} translation becomes a strictly unseen task.
As shown in Tab.~\ref{tab: flan 2021}, the instruction-tuned {\method}s demonstrate scalable zero-shot performance even without finetuning with German data, signifying that  {\method}s are able to follow instructions.


\paragraph{Extensive zero-shot evaluation with large-scale instruction tuning.}
We then follow the recommended settings and conduct larger-scale instructions tuning on the full Flan 2022 Collection~\citep{longpre2023flan} and run extensive evaluations\footnote{We continue to evaluate on the \textbf{IWSLT14} dataset.
Besides, we also evaluate several datasets used in~\citet{chung2022scalingflan}. In detail,
\textbf{MMLU}~\citep{hendrycks2020mmlu} includes multiple-choice exam questions from 57 tasks covering humanities, social science, STEM, and more. 
\textbf{TyDiQA}~\citep{clark2020tydiqa} is an open-book question-answering benchmark across 8 typologically diverse languages}.
Following \citet{chung2022scalingflan}, we named our instruction-tuned checkpoints on Flan 2022 Collection as \texttt{Flan-XLM-R}.
The results in Fig.~\ref{fig: zero-shot} suggest that the \texttt{Flan-XLM-R} models are indeed general-purpose zero-shot learners, and their zero-shot performance substantially improves as the model scales. 
In particular, we highlight the results on IWSLT14.
The largest model, \texttt{Flan-XLM-R-XXL} even achieves a 30.90 zero-shot ScareBLEU score, only 2.4 below the performance of widely adopted supervised transformer baselines (33.30 as shown in Tab.~\ref{tab: flan 2021}).
This indicates the Flan-XLM-R models produce a very good language generation quality.

\subsubsection{{\method}s can learn in context}

\begin{wraptable}[8]{r}{0.6\linewidth}
\vspace{-8mm}
\centering
\caption{SacreBLEU of instruction-tuned {\method}s on IWSLT14 \textsc{De}$\rightarrow$\textsc{En} under oracle lengths with removed instruction, where the model can only figure out the objective of the task through demonstrations.}
\label{tab: in context}
\resizebox{\linewidth}{!}{
\begin{tabular}{lccc}
\toprule
\textbf{\method Arch.} & \textbf{w/o demonstration} & \textbf{w/ demonstration} \\ 
\midrule
\texttt{XLM-R-LARGE} (304M)  & 2.58 & \cellcolor{green!26}5.38 \\ 
\texttt{XLM-R-XL} (2.8B)    & 4.42 & \cellcolor{green!30}12.18 \\ 
\texttt{XLM-R-XXL} (9.7B)   & 4.16 & \cellcolor{green!40}19.16 \\ 
\bottomrule
\end{tabular}
}
\end{wraptable}
We also evaluate the in-context learning ability of the  {\method}s. 
We construct an experiment on IWSLT14 with removed instructions. 
In this case, the model can only rely on the demonstration to figure out the task objective.

The result supports that {\method}s can learn in context.
The model is unable to produce the desired outcome without prior knowledge of the task. 
However, when given a demonstration, it can learn to treat the task as a translation task, showing obvious performance improvement, which also scales with model sizes.





\subsection{Exploring Reasoning with {\method}s}
\label{sec: reasoning}

We are also interested in exploring the reasoning abilities of our {\method}s as it is a crucial emergent ability that distinguishes LLMs from the small ones~\citep{wei2022emergent,fu2023specializing}.
Understanding how these models develop reasoning capabilities could provide insights into their scalability, generalization, and potential applications in complex problem-solving tasks. Moreover, investigating their reasoning mechanisms may help bridge the gap between traditional autoregressive LLMs and diffusion-based architectures, shedding light on their respective strengths and limitations in various domains.
In this section, we will highlight our key findings and include detailed discussions. 

\subsubsection{Quantitative Results}
\begin{wrapfigure}[35]{r}{0.5\linewidth}
  \centering
  \vspace{-5mm}
  \includegraphics[width=0.9\linewidth]{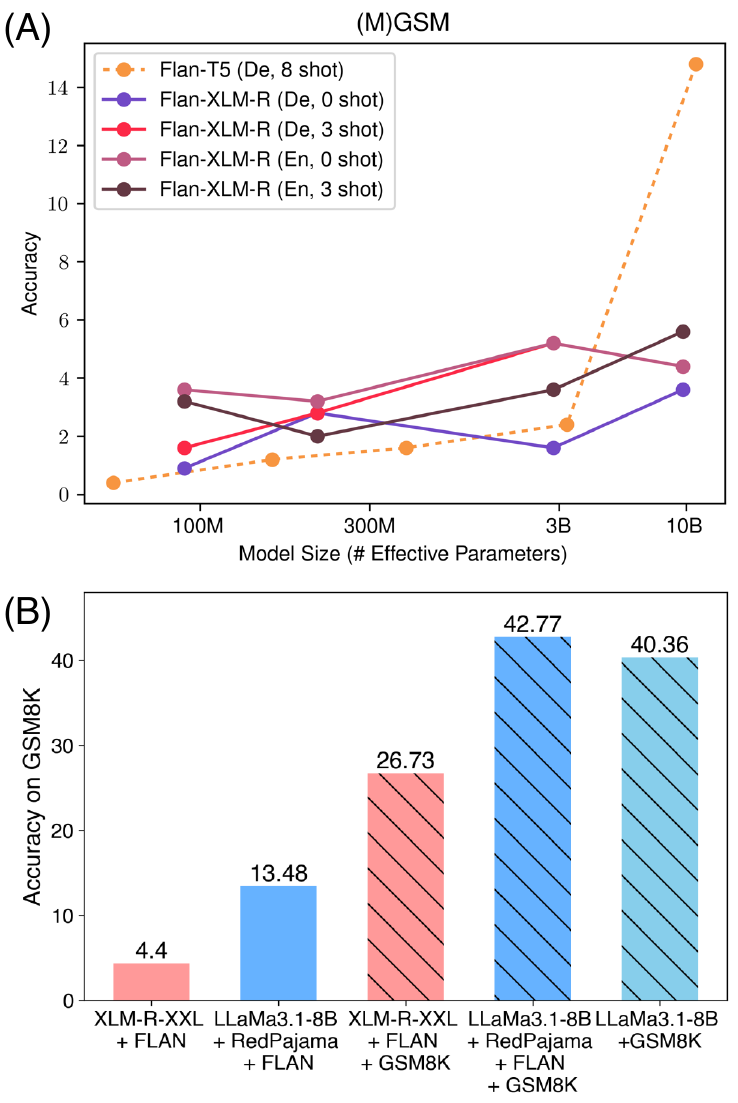}
  \vspace{-2mm}
  \caption{Evaluating reasoning on GSM datasets. (A) Performance of {\method} (\texttt{Flan-XLM-R}) and autoregressive LLMs (\texttt{Flan-T5}) of different scales. Although \texttt{Flan-XLM-R-XXL} fails to emerge with non-trivial performance, the performance improves with (B) task-specific tuning on GSM8K training set or better foundation model (adapted from LLaMa3.1 8B Instruct by continual training on RedPajama). We also include the result of fine-tuning autoregressive LLaMa3.1 8B on GSM8k for reference.}
  \label{fig: cot0}
\end{wrapfigure}

As shown in Fig.~\ref{fig: cot0}, we find, by simply instruction tuning, even \texttt{Flan-XLM-R-XXL} fails to emerge non-trivial reasoning performance on GSM8K~\citep{cobbe2021gsm}, a benchmark dataset for mathematical reasoning, and its German translated version in MGSM~\citep{shi2022mgsm}.

We conduct further analyses to understand the unsatisfying performance and conclude that this is due to the limitation of the pre-trained recipe instead of the {\method} paradigm.
\begin{itemize}[itemsep=2pt,topsep=0pt,parsep=1pt,leftmargin=10pt]
    \item \textbf{{\method} can perform reasoning tasks.} Given that task-specific fine-tuning offers an effective strategy to predict the existence of emergent ability~\citep{snell2024predicting}, we finetune \texttt{Flan-XLM-R-XXL} on the training set of GSM8K. We find that the model performance rockets to a non-trivial accuracy of \underline{26.73\%}, confirming the capability of {\method}s in reasoning.
    \item \textbf{Reasoning performance of 
    {\method}s can be improved with better pre-training recipes.} We suggest that the reasoning performance of \texttt{Flan-XLM-R-XXL} is limited by its pre-training recipe, which significantly falls short of modern designs~\citep{warner2024smarter}. As a preliminary verification, we adapt an LLaMa3.1 8B into {\method}s by fine-tuning on RedPajama\footnote{We adapt the autoregressive LLaMa into a {\method} by replacing the causal attentions with bidirectional ones and continue pre-train the model with diffusion objective. Meanwhile, we also post-process the output logits by right shifting for one token to fill the gap that the output of autoregressive models predicts the next token while the output of {\method}s predicts the token at the same positions as the mask tokens~\citep{gong2024scaling}.} and obtain 13.48 after instruction tuning on Flan 2022 and 42.77\footnote{This performance is comparable to fine-tuning autoregressive LLaMa 3.1 with GSM8K training set, which is 40.36. Although we note that fine-tuning on GSM8K actually degrades the performance LLaMa 3.1 8B from near 80, we consider the fine-tuning results can represent model performance with sufficient training under the training distribution and implies similar capabilities between autoregressive and {\method}s.} after task-specific fine-tuning on GSM8K. Both greatly outperform XLM-R-XXL with the same fine-tuning settings.
\end{itemize}
These findings support that {\method}s are also able to perform reasoning as autoregressive LMs do while we need an improved pre-trained models to fully unveil the potential.

\begin{figure*}[th!]
    \centering
    \includegraphics[width=0.95\linewidth]{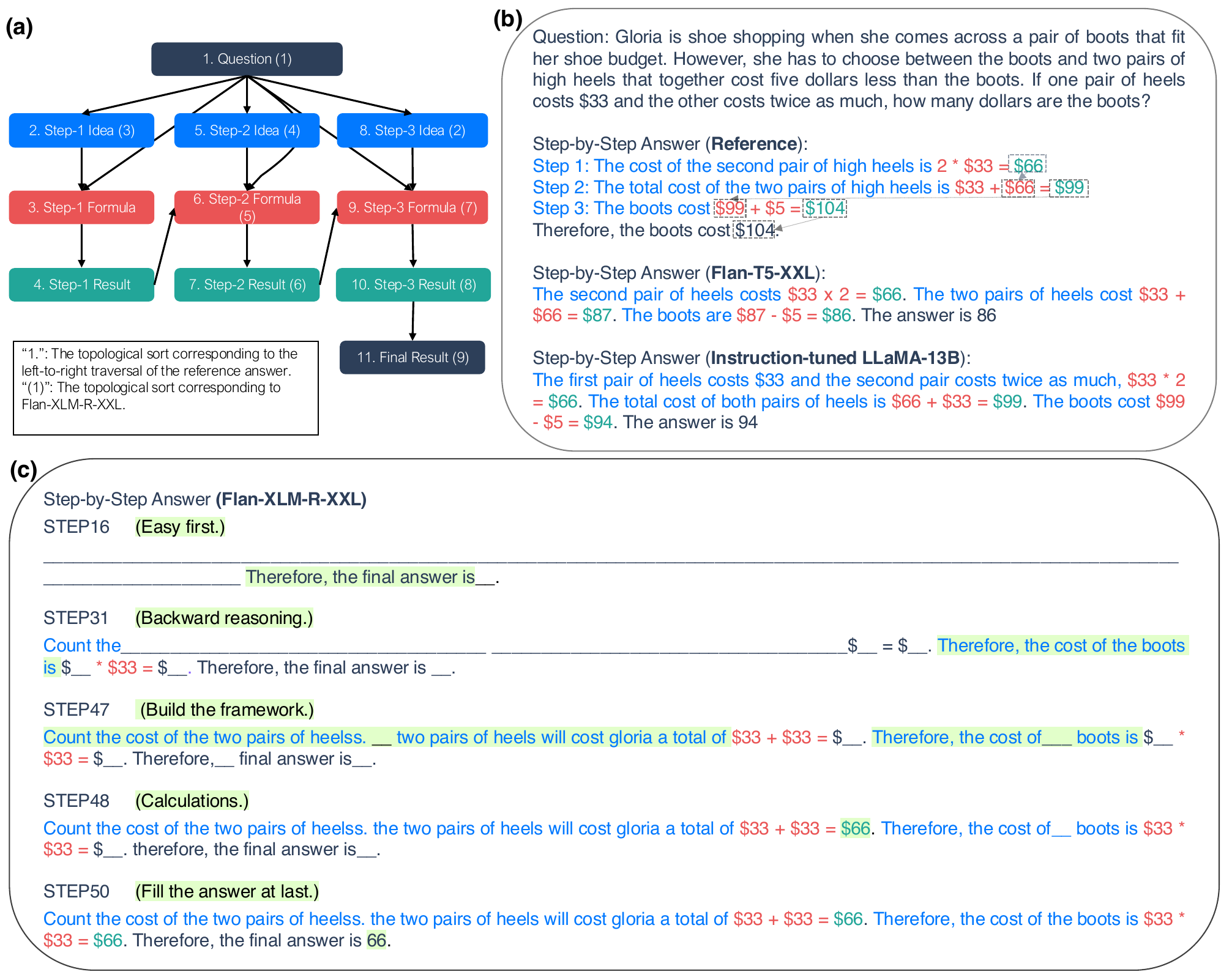}
    \caption{Qualitative investigation into the reasoning abilities of diffusion language models.
    \textbf{(a)} A causal graph~\citep{pearl1998graphical} that represents the dependencies between the reasoning steps. 
    \textbf{(b)} An example question, its reference answer, and answers from autoregressive models. 
    \textbf{(c)} The answer from \method (\texttt{Flan-XLM-R-XXL}) and its generation process.
    }
    \vspace{-5mm}
    \label{fig: reasoning}
\end{figure*}




\subsubsection{Qualitative Analysis}
\label{sec: qualitative}

In reasoning tasks, a model needs to generate intermediate reasoning steps to approach the final answers, where the model heavily relies on the intermediate results generated by itself to predict the final answer. 
This leads to constraints on the generation order when performing reasoning tasks.
Given the non-autoregressive nature of {\method}s, we wonder whether they show different behaviors in generation orders compared to fixed left-to-right autoregressive reasoning.




\paragraph{Understanding target dependencies with causal graphs.}
\label{sec: causal graph}
Fig.~\ref{fig: reasoning}(a) depicts the causal graph for the exemplary problem and its solution shown in Fig.~\ref{fig: reasoning}(b).
We argue that to solve the task with reasoning, language models must generate tokens in an order that conforms to a \textit{topological sort} of the causal graph.
Specifically, it means the following requirements for the generation order:
(1) the final results should come after the last intermediate result;
(2) the intermediate results should come after listing the corresponding equation;
(3) to correctly list an equation, models need to have the idea for this equation, copying calculation results from previous steps or numbers provided by the question; and 
(4) before these, models need to propose the idea for each step first.

\paragraph{{\method}s can figure out feasible topological sorts on the causal graph.}
A follow-up question is whether the generation process of autoregressive models and our diffusion language models conform to possible topological sorts.
One feasible topological sort is exactly the left-to-right traversal on the chain-of-thought text and is implicitly provided to autoregressive models during training.
Diffusion language models, on the other hand, learn without a fixed generation order due to random masking.
Fig.~\ref{fig: reasoning}(c) demonstrate its generation process of solving the exemplary question.
Despite incorrect final answers, the generative process does conform to a topological sort of the causal graph in Fig.~\ref{fig: reasoning}(a).
The model generates the ideas first, then writes the formulas, and finally calculates the answers.
We randomly sampled 30 samples generated by {\method}s and found that 21 out of these samples conformed to a topological order.
This implies that diffusion language models learn to figure out feasible topological sorts, namely a structure reasoning ability.


\paragraph{{\method}s reason with a flexible mind.}
Notably, diffusion language models are able to explore different topological sorts different from autoregressive models thanks to less constrained generative orders. 
We highlight some of the interesting patterns resulting from this.
\begin{itemize}[itemsep=2pt,topsep=0pt,parsep=0pt,leftmargin=20pt]
    \item \textbi{Easy first.} 
    Fig.~\ref{fig: reasoning}(c) shows that the model fills up the fixed pattern (\ie, ``the final answer is'') at first, showing a quite smart easy-to-hard generation behavior.
    
    \item \textbi{Planning ahead.} 
    In Fig.~\ref{fig: reasoning}(c), the model constructs the framework for the solution before diving into arithmetic. Actually, we have seen similar behavior in Fig.~\ref{fig: qualitative mt} where the model generates three clauses simultaneously. Both cases demonstrate the models' global perception which helps plan the generation of the whole sequence.
    
    \item \textbi{Forward and backward reasoning.} 
    During the reasoning process in Fig.~\ref{fig: qualitative mt}, on \underline{}{STEP 31}, the model begins the solution with the idea for the last reasoning step. This shows backward reasoning behavior, a very common human behavior that is especially helpful for challenging reasoning activities such as finding mathematical proofs~\citep{kazemi2022lambada}.
    
    \item \textbi{Backtracing.} 
    The backward transition of diffusion models formally supports backtracing by remasking. In Fig.~\ref{fig: reasoning}(c), \underline{STEP 47} erases a ``the'' token. This ability helps avoid accumulating errors in predicted tokens~\citep{arora2022exposure}. 
\end{itemize}

\subsubsection{{\method}s Show Superiority with Reasoning That Nessitates Implicit Planning}

\begin{wrapfigure}[21]{r}{0.5\linewidth}
  \centering
  \vspace{-5mm}
  \includegraphics[width=\linewidth]{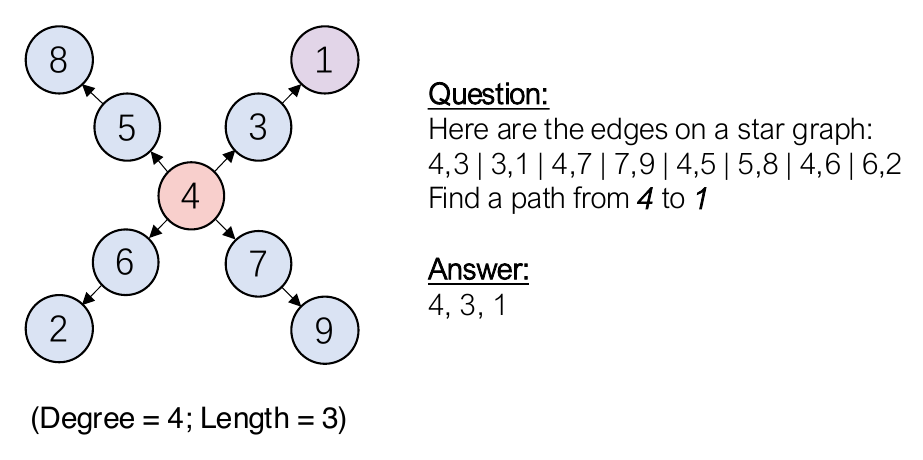}
  \vspace{-6mm}
  \caption{An example of Path-Finding on Path-Star Graph~\citep{bachmannpitfalls} with degree of central start as 4 and length of each path is 3. The task inputs the graph as well as the node indices of the central start (Node 4 in the example) and goal node (Node 1 in the example), and requires a model to figure out the path from the central start and the goal node (``Node 4, Node 3, Node 1" in the example). This task serves as a minimal planning task as it can be simply solved only if the models look ahead to see which first step leads to the goal.}
  \label{fig: star}
\end{wrapfigure}

Inspired by the qualitative observations, we consider that {\method}s could be helpful when the logical reasoning process differs from the sequential (\ie, left-to-right) order of words in the text.
This is common, especially for challenging reasoning tasks, implicit thought processes are required to plan before giving the outcome rationales, which is also known as meta-CoT~\citep{xiang2025towards}.

\paragraph{Path-Finding on Path-Star Graphs.}
To verify this, we experiment to see whether {\method}s are able to solve Path-Finding on Path-Star Graphs~\citep{bachmannpitfalls}, a simple planning-related problem that autoregressive models struggle with\footnote{We refer readers to \citet{ye2024beyond} which elaborate on the planning capability of \method in more details.}.
As shown in Fig.~\ref{fig: star}, a path-star graph contains a central start node with multiple paths of equal length radiating outward from it, where one of the paths leads to the designated goal node.
The task for the model herein is to find the correct path from start to goal.
The task is simple only if the model can look ahead to discover which first step leads to the goal.
The straightforward failure mode makes it representative of studying the planning capability of the models.

\begin{wrapfigure}[18]{r}{0.5\linewidth}
\vspace{-4mm}
  \centering
  \includegraphics[width=\linewidth]{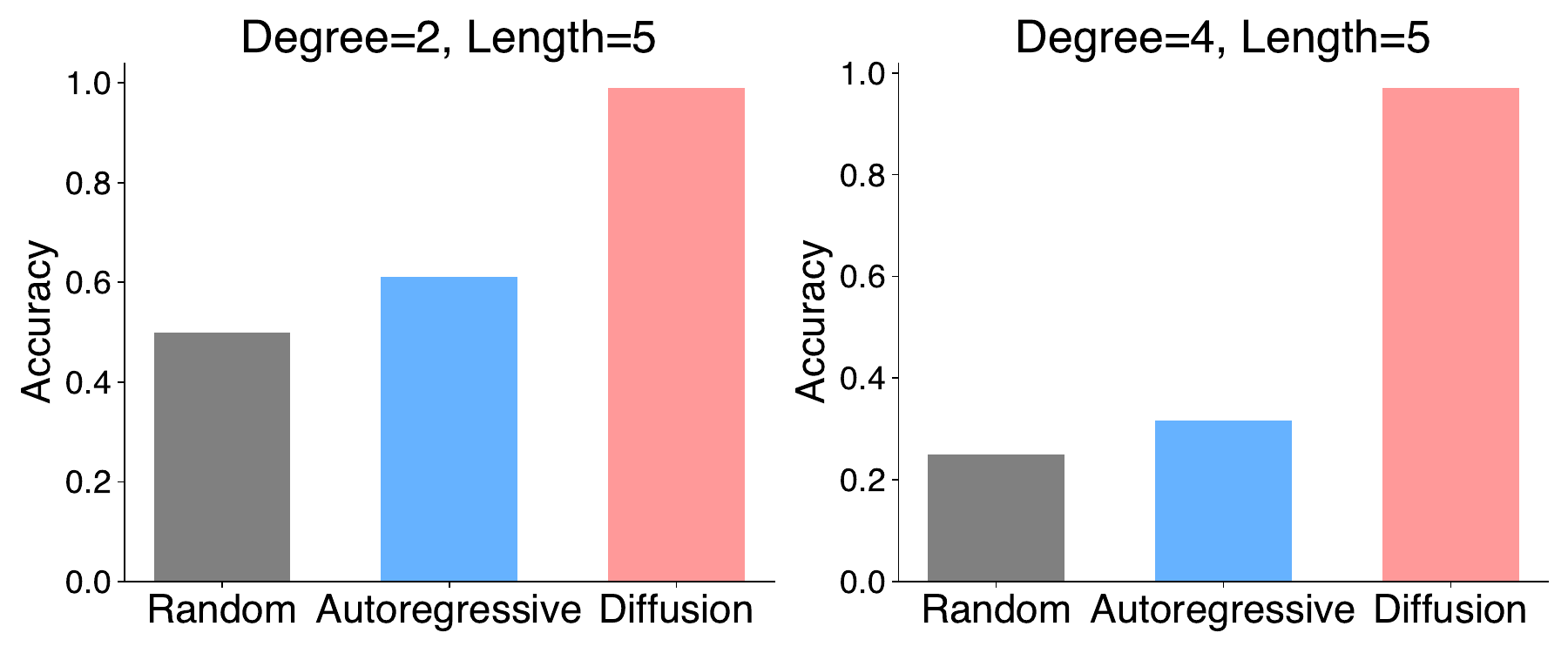}
  \vspace{-7mm}
  \caption{Comparison between autoregressive and {\method}s on Path-Finding on Path-Star Graph~\citep{bachmannpitfalls}.
  We experiment on two settings where the degree of the central start is 2 and 4, respectively.
  ``Random'': a random walk predictor that randomly selects a path on the star graph; ``Autoregressive'': fintuned \texttt{LLaMa3.1-8B-Instruct}; ``Diffusion'': {\method} fine-tuned from \texttt{Flan-XLM-R-XXL}.}
  \label{fig: star-result}
  \vspace{-8mm}
\end{wrapfigure}

\paragraph{Comparison with AR-LMs.}
The comparison between autoregressive models and {\method}s in Fig.~\ref{fig: star-result} reveals a significant advantage of {\method}s in solving reasoning tasks that require implicit planning. To investigate this further, we fine-tune two representative models—\texttt{LLaMa3.1-8B-Instruct} and \texttt{Flan-XLM-R-XXL}—on the Path-Finding on Path-Star Graphs tasks and evaluate their generalization performance on unseen graph structures.
Our evaluation encompasses two different experimental settings designed to test the models' ability to perform multi-step reasoning and implicit planning. In both cases, the autoregressive models exhibit limited success, struggling to capture the underlying graph structure and plan effectively. Their performance closely resembles a random walk, indicating an inability to leverage structural information for accurate predictions. In contrast, {\method}s consistently produce near-perfect predictions, demonstrating a remarkable capacity for handling implicit planning tasks.

This performance disparity highlights a key architectural difference: the bidirectional receptive field in {\method}s allows the model to capture global dependencies across the input more effectively. This not only facilitates better reasoning in structured environments but also gives {\method}s a clear advantage in tasks where planning and multi-step inference are required. These findings suggest that {\method}s are better equipped to model complex, non-sequential dependencies, offering new possibilities for reasoning-based applications beyond the capabilities of conventional autoregressive models.

\subsection{{\method}s as Multimodal Learners}

\begin{figure*}[t]
    \centering
    \includegraphics[width=\linewidth]{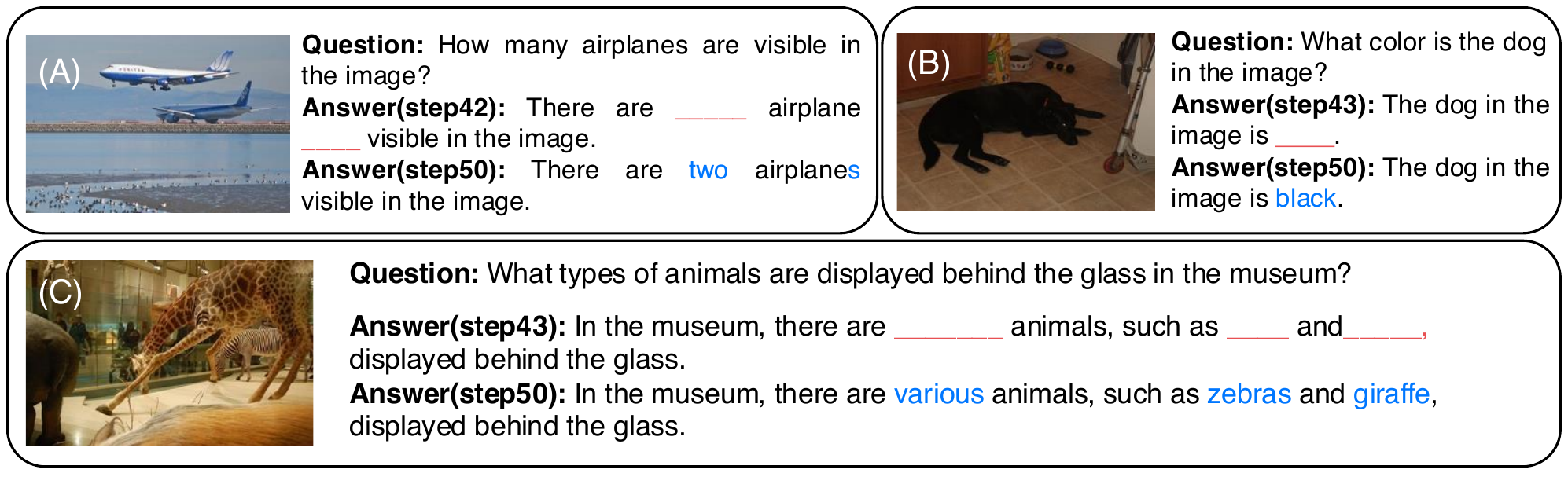}
    \caption{Visual question answering with {\method}s. We set the iterations steps for the {\method}s to generate final output as 50. Similar to the discussion in Sec.~\ref{sec: qualitative}, the model first generates easy content and fills the answer at the end.}
    \vspace{-6mm}
    \label{fig: vqa}
\end{figure*}
Recent advances in large language models extend beyond language processing, aiming to unify multiple modalities end to end for seamless multimodal interactions~\citep{zhan2024anygpt}.
Given the dominance of diffusion models in generating continuous signals~\cite{dhariwal2021diffusionbeatgans,bar2024lumiere} and excellent language capabilities shown in this study, we believe {\method}s contribute to a promising paradigm for developing unified multimodal models.
This motivates us to explore {\method}s for multimodal tasks.

In particular, we investigate whether {\method}s can tackle visual question answering (VQA).
We follow LLaVa~\citep{liu2024visual} to conduct two-phase training upon \textsc{Flan-XLM-R-XXL}.
In the first stage, we freeze the language model backbone and train a projector to map vision feature extracted from pre-trained CLIP visual encoder ViT-L/14~\citep{radford2021learning} to embeddings using the 558k subset of the LAION-CC-SBU~\citep{liu2024visual}. 
We then jointly tune the LM backbone and projectors for VQA with LLaVA-v1.5-mix665k data~\citep{liu2024visual}.

\begin{wraptable}{r}{0.5\linewidth}

\vspace{-6mm}
\small
\centering
\caption{Zero-shot exact match performance of \method and AR-LLMs on the dev set of GQA~\citep{hudson2019gqa}.}
\label{tab: gqa}
\resizebox{1.0\linewidth}{!}{
\begin{tabular}{lc}
\toprule
\textbf{Model} & \textbf{Exact Match} \\
\midrule
\method (\texttt{Flan-XLM-R-XXL})    & 39.93   \\
AR-LLM (\texttt{Flan-T5-XXL})       & 44.71 \\
\bottomrule
\end{tabular}
}
\end{wraptable}
Tab.~\ref{tab: gqa} shows the zero-shot performance of our models on the dev set of GQA~\citep{hudson2019gqa}.
For reference, we accordingly augment a \texttt{Flan-T5-XXL} model with vision understanding capability with the same recipe.
The results shows meaningful performance that supports the vision understanding capability of our model, which is close to that adapted from Flan-T5. 
Case study on Fig.~\ref{fig: vqa} shows that the model demonstrate similar behavior to what we find in the qualitative study for reasoning tasks (Sec.~\ref{sec: qualitative}) when generating answers for vision tasks.
In the three cases, the models answer in an easy-first order where they first generate content that can be copied from the question to build up the framework of the sentence and then fill in the key answers at the end. 
In Fig.~\ref{fig: vqa}(C), where the key answer contains multiple entities, {\method}s fill them simultaneously, showing the capabilities to parallelly process different vision information.

These results demonstrate {\method}s can also understand multimodal information similar to recent autoregressive LMs.
Together with the generation capabilities, for both well-tested vision generation~\citep{dhariwal2021diffusionbeatgans,rombach2022stablediffusion,bar2024lumiere} as well as language generation abilities verified in our study, diffusion models shed light on an appealing unified paradigm for multimodal foundation models.

\section{Discussions}
\label{sec: discussion}

In this work, we pioneer the study of the scalability of {\method}s to catch up with the recent advances of LLMs and facilitate the exploration of their potential. 
Our experiments verify their scalability regarding data, model sizes, and tasks.
Further, we showcase positive prospects about their reasoning capabilities such as casual-order generation and implicit planning for further exploitation.

\paragraph{Latest advancement on diffusion language models.}
After the first release of our study of \method, diffusion language models have attracted broad attention and plenty of closely related studies have emerged.
As such, we would like to highlight the latest progress to facilitate more upcoming advancements in this field.
\begin{itemize}[leftmargin=10pt]
    \item \textbf{Foundation.} To formulate a diffusion language model, \citet{ou2024your,sahoo2024simple,shi2024simplified,wang2024dplm,wang2024dplm2} also study masked language modeling-like objectives similar to ours and validate their effectiveness.
    Alternatively, \citet{loudiscrete} explores learning discrete diffusion language models by learning probability ratios as the extension of score matching and \citet{gat2025discrete} extends the flow matching formulation.
    All these studies confirm the practicality of building capable diffusion language models to serve as an alternative paradigm to autoregressive language models, with specific manners evolving. 
    \item \textbf{Scaling validation.} The scalability of diffusion language models under masked language modeling-like objectives has rapid progress.
    \citet{gong2024scaling} successfully build large-scale diffusion language models by adapting from autoregressive language models, offering another promising routine to gain large diffusion language models with relatively low cost.  
    \citet{nie2024scaling} investigate the pretraining of diffusion language models from scratch and show a scaling law parallel to autoregressive language models, indicating similar scaling trends of the two paradigms. 
    \citet{nie2025largelanguagediffusionmodels} further scales up the pretrained diffusion language models to 8B parameters and 2.3T pre-training tokens with up-to-date recipes, with results highlighting the competitiveness of diffusion language models to frontier open-source autoregressive models on well-recognized benchmarks for large language models and showcase a helpful chatbot built upon large diffusion language models.
    Besides natural language, \cite{wang2024dplm} scales diffusion language models to empower generative modeling of proteins.
    \item \textbf{Capabilities and applications.} 
    Diffusion language models have a bidirectional receptive field and can perform refinement by nature.
    For this reason, recent progress has confirmed that \textbf{\textit{diffusion language models show superiority in scenarios where left-to-right generation order is suboptimal}}.
    For instance, \citet{ye2024beyond} shows their advantage in tasks requiring implicit planning and \citet{nie2024scaling,nie2025largelanguagediffusionmodels} shows diffusion language models can fix the reversal curse of autoregressive models~\citep{berglund2023reversal}.
    Notably, the applications of diffusion language models demonstrate significant advances in scientific domains.
    With diffusion language models, \textsc{DPLM}~\citep{wang2024dplm} build frontier foundation models for protein and \textsc{DPLM-2}~\citep{wang2024dplm2} further investigate multimodal diffusion language models to unify generative modeling of protein sequences and structures.
\end{itemize}

We hope that our findings as well as our discussion on the latest progress can fuel the success of diffusion models in broader domains and also encourage investment into this compelling complement to autoregressive LLMs, which might push forward the boundary of techniques to pursue more advanced machine intelligence.





\bibliography{tacl2021}

\appendix

\onecolumn

\section{Implementation Details}
\label{app: details}
\subsection{Model}
Throughout this work, we mainly follow \citet{zheng2023reparameterized} to train and sample from our diffusion language models. 
Specifically, we set $\lambda_{t-1}^{(2)} = 1 - \frac{t-1}{T}$ in the training objective  where $t$ is the current timestep and $T$ is the number of total timesteps which is 50 in our experiments. 
Additionally, we apply label smoothing with a factor of 0.1 when we train a model without pretraining.
During sampling, we also follow \citet{ghazvininejad2019mask,savinov2021step,zheng2023LM_Design} and denoise tokens with high scores in each step instead of naively sampling from the Bernoulli distributions.
We use the same cosine schedule as in \citet{zheng2023reparameterized} to decide the number of denoised tokens in each step $k=\lfloor N\cdot\cos{\frac{\pi t}{2T}} \rfloor$, where $N$ is the sequence length.
For full details, we refer readers to the pseudocode in the original paper~\citep[][Algorithm 2]{zheng2023reparameterized}.
For length prediction, we feed model outputs into a one-layer transformer, apply mean pooling to the features and feed the pooled feature into an MLP classifier head.
For task-specific finetuning, we remove both input and output embeddings of the tokens that do not appear in the training set.

\subsection{Data}
For IWSLT14 and WMT14 machine translation tasks, we download and preprocess data following the example scripts provided by \texttt{Fairseq}\footnote{\url{https://github.com/facebookresearch/fairseq/tree/main/examples/translation}}, and we use SacreBleu~\citep{post2018call} for evaluation\footnote{The signature of sacrebleu for IWSLT14 \textsc{De}$\rightarrow$\textsc{En} is \texttt{nrefs:1|case:mixed|eff:no|tok:13a|
smooth:exp|version:2.3.1}, and for WMT14 \textsc{En}$\rightarrow$\textsc{De} \texttt{nrefs:1|case:mixed|eff:no|tok:intl|
smooth:exp|version:2.3.1}, respectively.}.
And we download Gigaword-10K data from the repository of LGEB\footnote{\url{https://github.com/CLUEbenchmark/LGEB}}.
For (M)GSM, we follow the instruction\footnote{\url{https://github.com/google-research/url-nlp/tree/main/mgsm}} in the official repository of \citet{shi2022mgsm} to process the data and prompts.
Besides, we obtain the preprocessed Flan 2021\footnote{\url{https://huggingface.co/datasets/Muennighoff/flan}}, Flan 2022\footnote{\url{https://huggingface.co/datasets/SirNeural/flan_v2}}, MMLU\footnote{\url{https://huggingface.co/datasets/cais/mmlu}}, 
and TydiQA\footnote{\url{https://huggingface.co/datasets/khalidalt/tydiqa-goldp}} from shared datasets on HuggingFace\footnote{\url{https://huggingface.co/datasets}}.
During training with Flan 2022, we follow the recommended ratios in \citet{chung2022scalingflan} to sample training data from different subsets.
We follow \citet{chung2022scalingflan} to report the MMLU performance on the validation set and adopt the GoldP setting for TyDiQA as in \citet{chowdhery2022palm,chung2022scalingflan}.
On the few-shot settings, we randomly select demonstrations. 
We will also release our code and data for better reproducibility.

\subsection{Training details}
We use Adam optimizer~\citep{adam} throughout our study. 
The dropout rate is consistent with the original configuration of the models which is 0.1.
For task-specific tuning, we use 8 Nvidia A100 GPUs. 
For instruction tuning, we use 8 Nvidia V100 GPUs for \textsc{BASE} and \textsc{LARGE}-sized models, 32 for \textsc{XL}, and 64 for \textsc{XXL}.
The overall batch size and other detailed hyperparameters for the two settings are in Tab.~\ref{tab: hyperparam task-specific} and Tab.~\ref{tab: hyperparam instruct}, respectively.


\begin{table}[h]
\centering
\small
\caption{The training hyperparameters for task-specific finetuning.}
\label{tab: hyperparam task-specific}
\begin{tabular}{@{}clccr@{}}
\toprule
\textbf{Dataset}                       & \textbf{Pretrained model} & \textbf{Batch size (\#. tokens)} & \textbf{Learning rate} & \textbf{\#. training steps} \\ \midrule
\multirow{4}{*}{IWSLT14 \textsc{De}$\rightarrow$\textsc{En}} & \texttt{XLM-R-BASE}       & 32K                 & 5e-5          & 150,000             \\
                              & \texttt{XLM-R-LARGE}      & 32K                 & 5e-5          & 150,000             \\
                              & \texttt{XLM-R-XL}         & 32K                 & 5e-5          & 100,000             \\
                              & \texttt{XLM-R-XXL}        & 32K                 & 5e-5          & 30,000              \\ \midrule
\multirow{4}{*}{WMT14 \textsc{En}$\rightarrow$\textsc{De}}   & \texttt{XLM-R-BASE}       & 128K                & 5e-5          & 300,000             \\
                              & \texttt{XLM-R-LARGE}      & 128K                & 5e-5          & 300,000             \\
                              & \texttt{XLM-R-XL}         & 128K                & 5e-5          & 150,000             \\
                              & \texttt{XLM-R-XXL}        & 128K                & 5e-5          & 100,000             \\ \midrule
\multirow{4}{*}{Gigaword-10K} & \texttt{XLM-R-BASE}       & 16K                 & 5e-5          & 30,000              \\
                              & \texttt{XLM-R-LARGE}      & 16K                 & 5e-5          & 10,000              \\
                              & \texttt{XLM-R-XL}         & 16K                 & 5e-5          & 5,000               \\
                              & \texttt{XLM-R-XXL}        & 16K                 & 5e-5          & 1,000               \\ \bottomrule
\end{tabular}
\end{table}


\begin{table}[h]
\centering
\small
\caption{The training hyperparameters for instruction finetuning.}
\label{tab: hyperparam instruct}
\begin{tabular}{@{}clccr@{}}
\toprule
\textbf{Training data}              & \textbf{Pretrained model} & \textbf{Batch size (\#. sequence)} & \textbf{Learning rate} & \textbf{\#. training steps} \\ \midrule
\multirow{4}{*}{Flan 2021} & \texttt{XLM-R-BASE}       & 512                       & 5e-5          & 5,000              \\
                           & \texttt{XLM-R-LARGE}      & 512                       & 5e-5          & 5,000              \\
                           & \texttt{XLM-R-XL}         & 512                       & 5e-5          & 3,000              \\
                           & \texttt{XLM-R-XXL}        & 256                       & 5e-5          & 1,000               \\ \midrule
\multirow{4}{*}{Flan 2022} & \texttt{XLM-R-BASE}       & 512                       & 1e-5          & 70,000             \\
                           & \texttt{XLM-R-LARGE}      & 512                       & 1e-5          & 30,000             \\
                           & \texttt{XLM-R-XL}         & 1024                      & 1e-5          & 17,000             \\
                           & \texttt{XLM-R-XXL}        & 2048                      & 1e-5          & 4,000              \\ \bottomrule
\end{tabular}
\end{table}
\clearpage

\newpage
\section{Full Experimental Results}
\label{app: full results}
The experimental results for task-specific tuning and instruction tuning on Flan 2022 are in Tab.~\ref{tab: full exp task-specific} and Tab.~\ref{tab: full exp instruct}, respectively.

~

~

\begin{table}[h]
\centering
\small
\caption{Full experimental results of task-specific finetuning. OL: the results are obtained with oracle length. LB: the size of length beam for length prediction.}
\label{tab: full exp task-specific}
\begin{tabular}{@{}clcccc@{}}
\toprule
\textbf{Dataset (Metric)}                                                           & \textbf{Setting} & \textbf{\texttt{XLM-R-BASE}} & \textbf{\texttt{XLM-R-LARGE}} & \textbf{\texttt{XLM-R-XL}} & \textbf{\texttt{XLM-R-XXL}} \\ \midrule
\multirow{2}{*}{\begin{tabular}[c]{@{}c@{}}IWSLT14 \textsc{De}$\rightarrow$\textsc{En}\\ (SacreBLEU)\end{tabular}} & OL               & 35.78               & 38.84                & 40.11             & 40.65              \\
                                                                                    & LB=10            & 34.10               & 37.33                & 38.54             & 38.57              \\ \midrule
\multirow{2}{*}{\begin{tabular}[c]{@{}c@{}}WMT14 \textsc{En}$\rightarrow$\textsc{De}\\ (SacreBLEU)\end{tabular}}   & OL               & 26.65               & 30.22                & 30.91             & 32.81                   \\
                                                                                    & LB=10            & 26.72               & 29.04                & 30.23             & 30.34               \\ \midrule
\multirow{2}{*}{\begin{tabular}[c]{@{}c@{}}Gigaword-10K\\ (Rouge-L)\end{tabular}}   & OL               & 28.83               & 31.33                & 31.72             & 32.57              \\
                                                                                    & LB=10            & 27.52               & 30.11                & 31.42             & 31.54              \\ \bottomrule
\end{tabular}
\end{table}

~

~

\begin{table}[h]
\centering
\small
\caption{Full experimental results of instruction tuning on Flan 2022. OL: the results are obtained with oracle length. LB: the size of length beam for length prediction.}
\label{tab: full exp instruct}
\begin{tabular}{@{}clcccc@{}}
\toprule
\textbf{Dataset (Metric)}                                                                    & \textbf{Setting}       & \textbf{\texttt{XLM-R-BASE}} & \textbf{\texttt{XLM-R-LARGE}} & \textbf{\texttt{XLM-R-XL}} & \textbf{\texttt{XLM-R-XXL}} \\ \midrule
\multirow{4}{*}{\begin{tabular}[c]{@{}c@{}}IWSLT14 \textsc{De}$\rightarrow$\textsc{En}\\ (SacreBLEU)\end{tabular}} & 0-shot (OL)   & 21.26      & 25.24       & 28.13    & 29.59     \\
                                                                                    & 2-shot (OL)   & 20.97      & 25.70       & 29.19    & 30.31     \\
                                                                                    & 0-shot (LB=3) & 17.76      & 25.12       & 26.42    & 30.90     \\
                                                                                    & 2-shot (LB=3) & 15.91      & 23.49       & 27.29    & 31.04     \\ \midrule
\multirow{2}{*}{\begin{tabular}[c]{@{}c@{}}MMLU\\ (Accuracy\%)\end{tabular}}        & 0-shot        & 31.28      & 32.79       & 40.17    & 42.13     \\
                                                                                    & 2-shot        & 28.74      & 32.72       & 38.08    & 42.06     \\ \midrule
\multirow{4}{*}{\begin{tabular}[c]{@{}c@{}}TyDiQA\\ (Rouge)\end{tabular}}     & 0-shot (OL)   & 23.42    & 62.28      & 83.39    & 84.76     \\
                                                                                    & 1-shot (OL)   & 23.46    & 66.86       & 86.12    & 84.36     \\
                                                                                    & 0-shot (LB=3) & 13.54      & 62.28        & 83.39    & 84.76     \\
                                                                                    & 1-shot (LB=3) & 12.63      & 44.49       & 55.78    & 84.36     \\ \midrule
\multirow{2}{*}{\begin{tabular}[c]{@{}c@{}}MGSM (\textsc{De})\\ (Accuracy\%)\end{tabular}}   & 0-shot        & 0.9        & 2.8         & 1.6      & 3.6       \\
                                                                                    & 3-shot        & 1.6        & 2.8         & 5.2      & 4.4       \\ \midrule
\multirow{2}{*}{\begin{tabular}[c]{@{}c@{}}GSM8K\\ (Accuracy\%)\end{tabular}}       & 0-shot        & 3.6        & 3.2         & 5.2      & 4.4       \\
                                                                                    & 3-shot        & 3.2        & 2.0         & 3.6      & 5.6       \\ \bottomrule
\end{tabular}
\end{table}

\begin{figure*}[t]
  \centering
  \includegraphics[width=\linewidth]{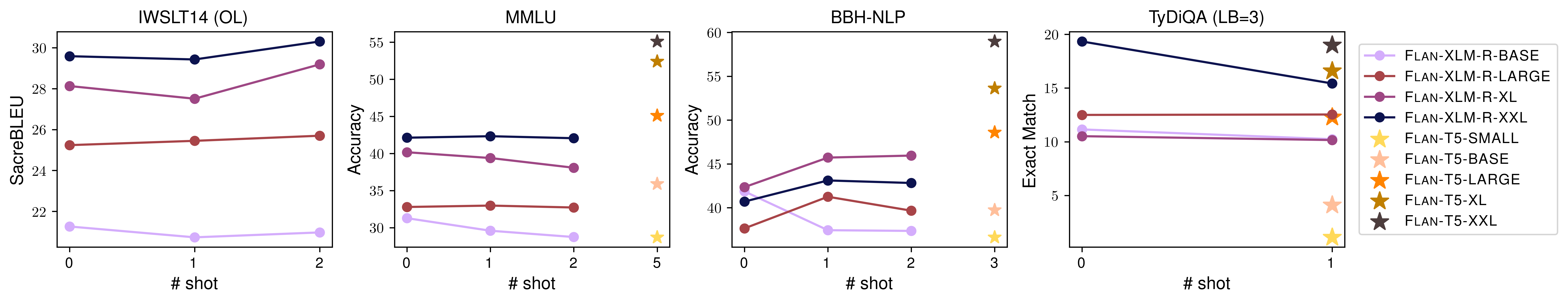}
  \vspace{-5.5mm}
  \caption{Few-shot performance of \texttt{Flan-XLM-R} and \texttt{Flan-T5} models. ``OL'' means the results are obtained with oracle length, while ``LB'' means the number of length beams to sample the target with length prediction. The model sizes refer to the number of non-embedding parameters.}
  \vspace{-1mm}
  \label{fig: few-shot}
\end{figure*}

\end{document}